# Kencorpus: A Kenyan Language Corpus of Swahili, Dholuo and Luhya for Natural Language Processing Tasks

*Wanjawa, Barack\**
    University of Nairobi, Kenya
    wanjawawb@gmail.com
*Wanzare, Lilian*
    Maseno University, Kenya
*Indede, Florence*
    Maseno University, Kenya
*McOnyango, Owen*
    Maseno University, Kenya
*Ombui, Edward*
    Africa Nazarene University, Kenya
*Muchemi, Lawrence*
    University of Nairobi, Kenya

**Abstract**

Indigenous African languages are categorized as under-served in Natural Language Processing. They therefore experience poor digital inclusivity and information access. The processing challenge with such languages has been how to use machine learning and deep learning models without the requisite data. The Kencorpus project intends to bridge this gap by collecting and storing text and speech data that is good enough for data-driven solutions in applications such as machine translation, question answering and transcription in multilingual communities. The Kencorpus dataset is a text and speech corpus for three languages predominantly spoken in Kenya: Swahili, Dholuo and Luhya (three dialects of Lumarachi, Lulogooli and Lubukusu). Data collection was done by researchers who were deployed to the various data collection sources such as communities, schools, media, and publishers. The Kencorpus' dataset has a collection of 5,594 items, being 4,442 texts of 5.6 million words and 1,152 speech files worth 177 hours. Based on this data, other datasets were also developed such as Part of Speech tagging sets for Dholuo and the Luhya dialects of 50,000 and 93,000 words tagged respectively. We developed 7,537 Question-Answer pairs from 1,445 Swahili texts and also created a text translation set of 13,400 sentences from Dholuo and Luhya into Swahili. The datasets are useful for downstream machine learning tasks such as model training and translation. Additionally, we developed two proof of concept systems: for Kiswahili speech-to-text and a machine learning system for Question Answering task. These proofs provided results of a performance of 18.87% word error rate for the former, and 80% Exact Match (EM) for the latter system. These initial results give great promise to the usability of Kencorpus to the machine learning community. Kencorpus is one of few public domain corpora for these three low resource languages and forms a basis of learning and sharing experiences for similar works especially for low resource languages. Challenges in developing the corpus included deficiencies in the data sources, data cleaning challenges, relatively short project timelines and the Coronavirus disease (COVID-19) pandemic that restricted movement and hence the ability to get the data in a timely manner.

## 1 Introduction

The intention to specifically focus research initiatives on low resource languages in Africa and other regions of the world is a necessity if we desire to preserve these languages. Language serves as a tool for both cultural preservation and communication. Language can also be used to gauge a community's success in terms of its economic, emotional, and social development (Smith, 2019). The advancement of natural language processing (NLP) in information technology, as used in machine learning and deep learning, has led to the creation of numerous useful applications such as text-to-speech, speech-to-text, machine translation, virtual assistants, text summarization, autocorrection and sentiment analysis, among others. However, these machine learning algorithms need training data, which is typically not available for low resource languages. Developing language datasets for languages with limited resources is therefore an appropriate initial

*\*corresponding author*

step to guarantee that machine learning tasks are feasible.

It is for this reason that the Kencorpus project was undertaken with the aim of collecting text and speech data for low resource languages. Nonetheless, we face the challenge of dealing with the many low resource languages of the world. For example, Africa has many different languages spoken within and across its borders. There are over 2,000 different languages spoken in Africa, representing a significant part of the world's languages (Heine & Nurse, 2000). A country such as Kenya alone has over 42 distinct language communities (National Museums of Kenya, n.d.). Therefore, the project started with a case study involving the three Kenyan languages of Swahili, Dholuo and Luhya.

The choice of these three Kenyan languages is based on purposive selection on relative representativeness. Swahili, also known as Kiswahili, is the national and official language of Kenya and Tanzania. The language is a cross border language in the Eastern part of Africa and is spoken by over 150 million speakers globally. Dholuo is a Nilotic language with an ethnic community of over 5 million speakers mainly around Lake Victoria in three East African countries of Kenya, Uganda, and Tanzania (National Museums of Kenya, n.d.; Omondi, 2020). Luhya, on the other hand, is a Bantu language of about 7 million speakers also predominantly used in the Western part of Kenya. It is a language with 17 sub-linguistic dialects i.e., Lulogooli, Luisukha, Luitakho, Lutiriki, Lubukusu, Lutachoni, Lunyore, Lumarachi, Lukhayo, Lusamia, Lunyala, Lumarama, Lushisa, Luwanga, Lutura, Lutsotso and Lukabras (Lubangah, 2018).

This research expects to develop and discover methodologies for collection, storage, and processing of corpora for under-resourced languages. The research also aims at developing datasets that are based on the corpus, such as Part of Speech (POS) annotation, translation across languages, Question Answering (QA) dataset and speech-to-text modeling for these low-resource languages.

The data presented in this paper is available for diverse machine learning data driven solutions such as question answering, machine translation and transcription. Through the project, we got insights into what it takes to prepare data (text and speech) and the accompanying datasets of part of speech (POS) annotations and translations between the low-resource languages. We have also developed a question answering dataset. Appreciating that such a corpora and datasets open many potential opportunities in the machine learning communities, the paper reports on benchmark work done on two use cases i.e. QA and speech-to-text (STT) Transcription.

The machine learning community involved in aspects of research that use corpora and datasets such as ours shall be the immediate beneficiaries. Enterprises interested in human language technology (HLT) systems can also access datasets for developing and testing their language models as they develop practical information and communication technology (ICT) systems e.g., chatbots, searching tools, translation systems, teaching aids, etc.

The rest of the paper is organized as follows – Section 2 provides the related work for this research while Section 3 describes the details of our methodology. Our results are presented in Section 4, while Section 5 discusses these results. Finally, Section 6 provides the conclusion and points out the areas of further research.

## 2 Related work

There are different issues to consider when conducting data collection and the curating of language datasets. These include the data collection process and the source materials for the language data that should be included in the corpus. There is also the need to consider the annotations that can result out of the corpus e.g., POS tagging, question-answer, among others. In addition, to show the viability of the collected datasets, researchers can develop benchmark models as proof-of-concept systems. In this section, we discuss related work based on the 3 areas that can be considered when developing language datasets. We focus on data collection with reference to building datasets for low-resource languages.

## 2.1 Data collection and African low-resource language corpora

Data collection modalities need careful consideration and planning. Participatory methods of data collection is a workable method that achieve a sustainable data collection model (Nekoto et al., 2020). As an example, some work has been done on using community initiatives to build corpora for low resource African languages. Such initiatives include the Masakhane project (Orife et al., 2020) and AI4D African language program (Siminyu et al., 2021), that focus on curation of language datasets. Another example of a participatory approach in data collection is the Digital Umuganda corpora in the country of Rwanda (Digital Umuganda, n.d.) where communities gathered at centralized locations for data collection activities. Other corpora developed for African languages have exploited the task of machine translation. This includes work done on building machine translation (MT) models with 204 languages, including some low resource languages of Africa (Costa-jussà et al., 2022). The Autshumato dataset is also a corpus for machine translation of the official languages of South Africa, though it can be useful for any other translations between pairs of languages (Groenewald & Fourie, 2009).

Open-source data can also be used for corpora compilation. Tracey et al., (2019) built a dataset for low resource languages including some African languages like Kiswahili, Wolof, Zulu, among others. They built the corpora through web-scraping and employed native speakers to search the web for texts in their native languages. Babirye et al. (2022) curated texts and speech datasets for 5 East African languages, including Kiswahili and 4 languages spoken in the country of Uganda. They employed both participatory approaches with local communities and web scraping of social media, Wikipedia and local news sites. Marivate et al. (2020) created low resource datasets for South African languages (Setswana and Sepedi) through scraping of online news channels. Apart from news sites, other resources contain collections of stories on African languages. These include the African story books (African Story Book, n.d.) and Tuvute Pamoja initiative (Tuvute Pamoja, n.d.) both of which have story collections in different African languages, while Edutab (Edutab, n.d.) provides a collection of Swahili story texts. Other datasets such as named entity recognition (NER) systems for ten African languages have been developed as part of uplifting low resource languages of Africa (Adelani et al., 2021).

Datasets of news articles in Kinyarwanda and Kirundi languages of Eastern Africa have been compiled for purposes of text classification and cross-lingual learning (Niyongabo et al., 2020). Web datamining is a method that has been used to develop some corpora such as the WebCrawl African where 694K parallel sentences for English and 15 African languages, such as Swahili, Zulu, Hausa among others, is done. Such data is applicable in machine translation tasks (Vegi et al., 2022).

For the Swahili language, the Helsinki corpus of Swahili (Hurskainen, 2004a) and the Swahili language online part of speech tagging tool Swatag (aflat, 2020) are tools for Swahili language processing, specifically Part of Speech (POS) tagging.

Kenya languages with some resources include the Kikuyu language which has a spell checker utility (Chege et al., 2010). There are few corpora for the Kenyan low resource languages of Dholuo and Luhya. Nonetheless, there exist some in the public domain. Some Dholuo texts have previously been collected for the task of machine translation from religious texts, though it can be used for other machine learning tasks (de Pauw et al., 2010).

For the Luhya dialects, Siminyu et al. (2021) developed a phone dataset for Lubukusu and Lusaamia dialects. Ngoni (2022) developed an English-Luhya machine translation dataset for Lubukusu dialect of Luhya, while Chimoto & Bassett (2022) developed an English-Luhya machine translation dataset from Bible translation for the Lumarama dialect.

Speech data can be collected using methods such as GroupTalk where data collection is done through interviews in small groups of focused group discussions in formal settings (Cieri et al., 2002). Conversion of written text to create speech data, such as the use of GroupMeet has been used previously to extend data corpora (Gelas et al., 2012). Mukiibi et al. (2022) leveraged on local radio stations to collect speech data for Luganda, spoken in Uganda.

These tried and tested methods of using a combination of participatory approaches, web-scraping and local radio stations have been found useful and have been adopted in developing the Kencorpus datasets. These research efforts have led to a collection of language corpora and datasets, but there is still more that needs to be done to resource the many low resource languages of Africa and the world.

## 2.2 Creation of annotation datasets

For this research, we focused on three possible annotations that can be developed on top of the collected language datasets namely Part of Speech (POS) tagging, Question-Answer pairs for QA task and Parallel corpora for Machine Translation.

Part of speech (POS) tagging is usually one of the annotation datasets that are needed in an NLP pipeline. POS tagging is therefore an important aspect of data curation that researchers of low-resource languages should consider. Some open-source toolkits exist for POS tagging e.g., GATE (Cunningham, 2002), and BRAT (Stenetorp et al., 2012). These are applicable for high resource languages but can also be tweaked for low resource languages (Skeppstedt et al., 2016). However, these toolkits may not be practical in all cases, especially where there is no internet access during annotation. Annotators also need some training in the use of these tools. In such scenarios and as an alternative, POS annotation can be done using spreadsheets as was done for some low resource languages of Kenya such as Kikamba (Kituku et al., 2015; Pauw et al., 2006). In this research, we adopted the use of spreadsheets as the project duration was short and it would have taken longer to train the linguistic annotators on the use of a toolkit such as GATE (Cunningham, 2002).

POS annotations require tagsets. Different languages tend to have different tags e.g., the Kiswahili tagset (Hurskainen, 2016), Kamba tagset (Kituku et al., 2015). This realization therefore leads to the need for some middle ground tagset, such as the universal target set proposed by Petrov et al. (2011). For this work, we adopted the POS tagset proposed by Petrov et al. (2011), as shown in Appendix 1, to enable uniform deployment to the three different languages considered in this research.

Another necessary annotation dataset that has gained popularity among low resource language researchers is a parallel corpus for machine translation. Toolkits for neural machine translation already exist, such as the openNMT toolkit which uses neural networks to perform the translation (OpenNMT, n.d.). Some work has been done on machine translation such as neural machine translation models used across 5 different languages in South Africa with 50,000 sentences (Martinus & Abbott, 2019). Practical applications of translations for low resource languages include work done in translating a glossary of Coronavirus disease (COVID-19) terms across 33 languages (Translators Without Borders, n.d.).

Initiatives to develop machine translation corpora for low resource languages include work done on Bambara language of West Africa to and from English and French (Tapo et al., 2020). Statistical and neural translation methodologies have also been tried for Somali and Swahili languages spoken in Eastern Africa (Duh et al., 2020). Chimoto et al. (2022) creates a parallel corpus of 8,000 Luhya-English sentences for purposes of machine translation. These works bring out the challenges and opportunities of working with languages that have little data sources.

Finally, a QA dataset is one of the datasets that can be developed from a language corpus. Several QA datasets exist for high resource languages e.g., SQuAD (Rajpurkar et al., 2016), MCTest (Richardson et al., 2013), Common sense knowledge systems (Ostermann et al., 2018), WikiQA (Yang et al., 2015), TREC-QA (Voorhees & Tice, 2000) and TyDiQA (Clark et al., 2020). However, only a few QA datasets are available for low-resource languages, in particular Kiswahili. TyDiQA is such a dataset since it has a QA collection of 11 languages from the Wikipedia corpus including the low-resource language of Swahili (Clark et al., 2020). It is therefore desirable to deliberately develop more QA datasets, especially for low-resource languages.

## 2.3 Proof of concept

Proof of concept systems are important in testing and confirming the useability of a corpus and its resultant annotation datasets. It is possible to test different NLP aspects of a corpus and dataset e.g., effectiveness of the POS tags in NLP tasks, practical use of the translation sets in NLP systems, and even benchmarking the QA set against QA-related tasks.

For example, assessing the effectiveness of a QA dataset can be done using machine learning. This is however only possible when there exist vast amounts of data to train, validate and test the models. Use of machine learning QA models on datasets such as the English language SQuAD QA dataset (Rajpurkar et al., 2016) has achieved F1 scores of up to 93.214 (Paperswithcode, n.d.). That was achieved using bidirectional encoder representations from trans-former (BERT) ensembles, which is a deep learning method that relies on lots of training data. However, in cases where there is little or no training data, as is usually the case with low re-source languages, there is the need for other methods that may not require as much training data. One such alternative data model uses semantic networks (SN) (Wanjawa & Muchemi, 2020, 2021). SNs are already used in domains such as Google Knowledge Graph (Singhal, 2012), LinkedIn (Wang et al., 2013) and Facebook (Sankar et al., 2013) amongst others. SNs can be formulated to model natural language data. Even in such cases, there is still need for some additional part of speech (POS) tagged corpus, to tag the source text, for this SN method to work. Once tagged, it is then possible to employ SNs to first model the language in such a way that the language is structured and ready for tasks such as QA.

SNs model the language by creating a network of triples having subject-predicate-object (SPO). If a relationship of these triples can exist, then a meaningful knowledge base can be created. Fortunately, languages also have a generic structure that gives them meaning. For example, Swahili has a generic subject-verb-object (SVO) structure (Hurskainen, 2004a; Ndung'u, 2015). It is therefore possible to map the SVO structure of a language, to the SPO structure of a SN. The only processing of the language text that is needed is therefore just the tagging of the source text with the corresponding POS. No other training data is needed, hence overcoming the need for training and testing data. The POS tags can be used to identify these subjects and objects (usually nouns) and predicates (usually verbs).

STT transcription systems are of practical use in applications that assist people with hearing challenges. These systems develop automatic speech recognition (ASR) systems that exploit speech data, such as that available in a typical corpus like this project's Kencorpus speech corpus. It is therefore possible to test ASR systems on the speech data of a corpus as proof of concept, especially for low resource languages where few such systems exist. Transcriptions usually follow a five-stage workflow in processing the speech files into text (Moore & Llompart, 2017). This is a pipeline that can be tried in a proof-of-concept system for a low resource language, together with associated toolkits such as CMU Sphinx to create and test an ASR system (Pantazoglou et al., 2018; Reddy et al., 2015).

## 3 Methodology

The Kencorpus project collected primary data, both speech and text, in the three Kenyan languages of Swahili, Dholuo and Luhya. The project then curated text and speech corpora for each individual language. Additionally, three datasets were developed from annotations done on the language corpus. The first additional dataset was a QA dataset for Swahili, the Kencorpus Swahili Question Answering (KenSwQuAD) dataset. The second was a set of translations of Dholuo and Luhya language texts into Swahili, while the third was a part of speech (POS) tagged Dholuo and Luhya texts. The project also developed proof of concept systems for STT modeling and QA system. The details of each aspect of the project are provided in this section.

### 3.1 Choice of languages

Kenya has more than forty-two indigenous languages and are grouped in three major language families of Bantu, Nilotic and Cushitic (Iraki, 2009). Kenyan languages tend to have predominance in certain geographical locations within the country. The choice of Swahili, also known as Kiswahili, as a language in the project was due to Swahili being the national language of Kenya. Despite it being spoken by many speakers in East Africa and with interests globally, Swahili is still a low resource language. More deliberate research efforts are therefore still need-ed to provide corpora and machine processing tools for Swahili. There are many dialects of Swahili (Wald & Gibson, 2018; Walsh, 2017). The Swahili data collected (text and speech) was mainly the Standard Swahili that is of general use in official and learning settings, though subtle differences were possible depending on the region of Kenya where the data came from. The Swahili data in this research is therefore considered as Standard Swahili.

Dholuo is a Nilotic language mainly spoken in the Western part of Kenya near Lake Victoria. It also has speakers in the countries of Tanzania and Uganda. It is the second most populous language in Kenya (Mazrui, 2012). Dholuo also has different dialects or sociolects, and the project collected language dialects that were available in the data collection field. Nonetheless, these dialects tend to have mutual intelligibility and hence can be generally called Dholuo.

The Luhya language is a Bantu language also spoken predominantly in the Western part of Kenya. This language however comprises seventeen different dialects within it (Lubangah, 2018), hence is a language of interest due to its diversity amongst its speakers. Due to the constraints in resources, only three dialects within the Luhya language were selected, being Lumarachi, Lulogooli and Lubukusu. The purposive sampling was guided by Lubukusu and Lulogooli being the most populous languages compared to the other dialects (Lubangah, 2018).

The Lumarachi dialect was of interest due to the geographical location of most of the speakers who are at the border of the Dholuo speakers in Western Kenya. The proximity was of interest to gauge if there was any cross-border effect in these languages of study. The researchers also had ease of access to the geographical location and internally available resources to interact with the Dholuo and Luhya languages chosen.

### 3.2 Scope of project

The Kencorpus project aimed at collecting speech and text data for the three languages of Swahili, Dholuo and Luhya. The intention was to collect an equal number of speech and text data in these three languages.

Additionally, the collected data was to be annotated with parts of speech (POS) for Dholuo and Luhya texts. Swahili texts were not tagged since existing projects have already developed POS tagging for Swahili which are generally dependable (Hurskainen, 2004a, 2004b). The third aspect of the project was to undertake translation of texts from the Dholuo and Luhya languages into Swahili. This was to increase the size of the Swahili corpus, while also understanding the intricacies of such translations. The datasets from translation would also be useful for the task of machine translation. A fourth component of Kencorpus was to create a QA dataset based on the Swahili text corpus. Finally, the project aimed to develop proof of concept systems to confirm that the collected data and annotations were of practical use to the machine learning community. To this end, two proof of concept systems were developed. These were a QA model for Swahili texts, and a STT model for Swahili speech files.

### 3.3 Data sources

The project identified data sources which were both primary and secondary data sources. The primary data was collected using participatory approaches from institutions of learning (schools, colleges), local community engagements and during social events. Secondary data was obtained from partnering media houses and publishers. Research assistants, who are natives of the respective languages, visited the geographical locations of the research areas and assisted in the collection of the data. The project employed three research assistants per language to lead data collection efforts. We had partnerships agreements with several institutions to enable us to access their existing datasets or to allow us to link to their data sources. This enabled us to collect texts from various genres such as articles, book sections, news texts, African short stories, and other publications. Additionally, the project held story writing competitions in educational institutions with the aim of getting texts from different geographical regions and language dialects. Our respondents were purposively drawn from different genders, age groups and geographical locations. We also collected web-scraped data from tweets particularly for Kiswahili and Dholuo.

### 3.4 Data collection

The project designed guidelines and methods for collecting both text and speech data from various sources. Text data was collected through story competitions mainly in schools. This meant providing the respondents with writing materials (pens, papers) and collecting the resultant creative writing, which were later digitized (see Section 3.6 for details). However, volunteers and college-level students provided their own compositions already in soft copy. Local media houses broadcasting in local languages (Dholuo, Luhya) provided their news transcripts in soft copy. Publishers provided their data either as hard copies for photocopying or in some cases as soft copies ready for direct processing by computers. Tweets (for Kiswahili and Dholuo) were collected using Twitter API (Lomborg & Bechmann, 2014) with search phrases that included names of known local radio stations and journalists who post regular content in the local languages. Any source material that was obtained as hardcopies were later digitized as detailed in Section 3.6.

Speech data was recorded using voice recorders (or voice recording apps on smartphones like Easy Voice recorder[1]), while collaborating media houses provided speech data through computer files mostly in MP3 format. All collected raw data was compiled into an initial database and classified as level 0 data.

In terms of Research Ethics, we developed research consent forms for use in the project. The consent form spelt out the project objectives and how the data would be processed, accessed, and used. Only the respondents who were willing to provide data were allowed to participate in the project, subject to their informed consents. All respondents who agreed to participate in the project signed the forms and also kept a copy of the consent forms. Consent forms for groups such as schools were executed by the school managers on behalf of the respondents. Even in such cases, only the students that were willing to participate in the project were allowed to do so. All members participating, whether individuals or groups, were listed by full names and their contacts on the consent forms. Metadata collection forms, which were hard copy data forms ready for population by pen, were also developed to capture the details of each data item collected. All these items were part of the research assistants' toolkit. The metadata details captured during the project are shown in Table 3.1 below.

*Table 3.1: Kencorpus project statistics*

| Aspect | Data to be provided |
|---|---|
| 1. Date | Date of data collection |
| 2. Story ID | Unique identification of the data item collected |
| 3. Researcher details | Needed: (1) Researcher Name (2) Resource Person Name |
| **4. Story details** | |
| 4.1 Language | Options: Swahili(Swa) / Dholuo(Dho) / Luhya- Lumarachi(Lhych) / Lubukusu(Lhybk) / Lulogooli(Lhylg) |
| 4.2 Category of story | Options: Letters / News / Folks / Essays / Story / Dialogue / Song / Advertisement / Others |
| 4.3 Topics | Options: Agriculture / Education / Culture / Commercial / Political / Religious / Social affairs / Others |
| 4.4 Title | Title of the data item being collected |

1 - https://play.google.com/store/apps/details?id=com.coffeebeanventures.easyvoicerecorder&hl=en&gl=US&pli=1

| 4.5 Short description | A description of the data item being collected |
|---|---|
| 4.6 Type | Options: Text / Speech |
| **5. Origin** | |
| 5.1 Location | Geographical location where data collection was done |
| 5.2 Source | Options: Schools / Publisher / Media / Community |
| 5.3 Name | Name of author / speaker / provider |
| 5.4 Gender | Options: Male / Female |
| 5.5 Age group | Options: School going / Community / under-18 / Adult / Elderly |
| 5.6 Contacts | Tel./email |
| **6 Consent form** | |
| 6.1 Was signed | Date / Confirm copyright owner given right for reproduction? |
| 6.2 Was not signed | Confirm considerations to ensure no copyright infringement |
| 6.3 Include in corpus | Options: Yes / No |
| **7. Text processing** | |
| 7.1 Pages/words | Number of pages on the text (or words) |
| 7.2 Text source | Options: handwritten / typed / already softcopy / image-scan |
| 7.3 Source text quality | Options: excellent / good / acceptable / bad-unclear |
| 7.4 Source text format (if softcopy) | Options: TXT / DOC/X / PDF / ODT / picture/scan |
| 7.5 Initial computer copy | Options: re-typing / scanning / already softcopy |
| 7.6 Conversion to text | Options: re-typing / scan to text OCR / already softcopy |
| **8. Speech processing** | |
| 8.1 Setting | Options: Studio / Clean (anechoic chamber) / Natural recording (quiet room) / Recording with background noise / Telephone recording |
| 8.1 Recorder | Options: Speech file as provided / Phone app details / Microphone details |

### 3.5 Staff Training and Piloting

The Kencorpus project undertook training of the identified research assistants, who were knowledgeable in the three research languages. This training was done at the beginning of the project. Both face to face and virtual training were done. Internal capacity building between researchers within the language categories was also done. Data collection efforts were expected to be evenly distributed amongst the three researchers per language or language dialect.

The researchers then started their field work while reporting back to the project office on their field experiences. Such experiences enabled the project to fine tune their research tools to be better suited to the research environment and continually improve them over time.

Subsequent training was done for data cleaning and annotation tasks. Data cleaning was aimed at converting the raw collected data at Level 0, into edited computer formats e.g., text images or scans were to be converted to the standard text (TXT) format. Any other text for-matted documents were also to be reformatted back to TXT. Speech files were all to be con-verted to the standard Waveform Audio File (WAV) format.

### 3.6 Data Cleaning

In this project we did data cleaning of texts only. The speech clips were retained as originally recorded due to constraints in equipment to edit speech files. However, original recorded speech files passed through quality control checks to assure the best quality possible. Most of the speech files were also studio quality, having been obtained from media houses.

Data cleaning of texts is an essential process of eliminating noisy signals that would other-wise degrade the quality of datasets intended for natural language processing. The noise in data can be due to the presence of corrupted characters, misspellings, inconsistent data, redundant data, missing characters, extra spaces, inconsistent punctuations, spelling variations, and codeswitching, among others. The data collected in this study was not devoid of these noisy elements. Therefore, a systematic approach was developed to ensure that the raw corpus comprising handwritten manuscripts, scanned pages, images, web-scraped data (tweets), and text files were adequately processed to deliver quality and clean texts in the language datasets.

The data cleaning approach adapted a four-stage data cleaning process (den Broeck et al., 2005). However, we tweaked it by introducing a preceding stage that we named 'Digitize', and also introduced a different processing route for tweets. The final data cleaning pipeline that we adopted was therefore a 5-stage process that included the digitization, screening, diagnosis, treatment, and documentation stages. The data cleaning process flow is shown in Fig. 1.

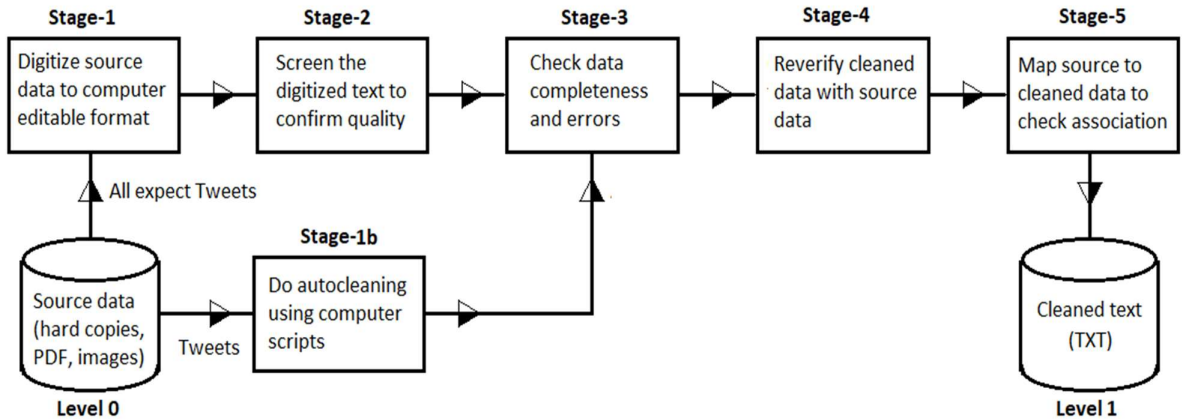

**Fig. 1** - The 5-stage data cleaning process adopted by Kencorpus project (source: author).

In the full data cleaning cycle, we employed eighteen data cleaners composed of native speakers from the three languages. They were interviewed, hired, and trained on data cleaning and research ethics. To ensure data reliability, each file had a primary and secondary data clean-er. As an initial process, the primary data cleaner meticulously went through each sentence in the documents to clean it according to the data cleaning scheme. The cleaning done included identification and removal of noisy elements, correction of spelling and grammatical errors using human language experts, and handling of missing data that could have been omitted in the process of conversion of source text to computer format, among others. As a second process, at the end of the week, the data cleaners were swapped within their respective language teams to look through already cleaned text files to recheck them for any previously unseen errors. They would then further correct such errors to improve the data quality. Finally, a linguist, for the respective languages, randomly selected and perused sample files for quality assurance of the data cleaning. The linguists were in charge of cleaning and also the other annotation tasks described in section 3.7.

The tweets were cleaned automatically using Python tweet-processor and regular expressions to remove constructs such as webpage uniform resource locators (URL), hashtags and emojis. Moreover, all the tweets were lowercased, and the images, and other sections of the tweet dropped to only retain the message section. The final processing entailed the use of a regular expression library to remove non-ASCII characters that could have been left behind. Fig. 2 below shows a sample of a processed tweet (in Kiswahili), with the raw scrapped tweet on the left, and the processed tweet on the right.

| Raw Tweet | Processed Tweet |
|---|---|
| Mwanamke Bomba Esther Muiu hujishughulisha na shughuli za kuwasaidia wanafunzi Mama huyu wa watoto wanne na wajukuu husaidia wale wasiojiweza Esther hushirikiana kwa karibu kuwasaidia wanafunzi wa kike #SemaNaCitizen https://t.co/oocALvHjwf | Mwanamke Bomba Esther Muiu hujishughulisha na shughuli za kuwasaidia wanafunzi. Mama huyu wa watoto wanne na wajukuu husaidia wale wasiojiweza Esther hushirikiana kwa karibu kuwasaidia wanafunzi wa kike |

**Fig. 2** - Processing of tweet from original TXT to final TXT for part of Kencorpus storyID 2597

The first step in processing texts that were not tweets, was to digitize the manuscripts by scanning and converting the scanned portable document format (PDF) images into an editable text format. Approximately three thousand manuscripts, largely composed of compositions written by primary and secondary school students, were scanned into PDF text images. Subsequently, several Optical Character Recognition (OCR) software were used to convert the PDF files to editable text formats, with mixed results: some achieving very accurate digitization while some were not very accurate. This could largely be attributed to the handwriting quality and partly to the precision of the OCR software. The OCR software used were Pen-to-Print mobile phone app (Pentoprint, 2021), Expert PDF OCR (Avanquest Software, 2021), and Google doc internal file converter. An example of an original handwritten text page subjected to OCR processing using Pen-to-Print is shown in Fig. 3. The original handwritten text scan is shown on the left side of the figure, while the processed text is shown on the right-hand side.

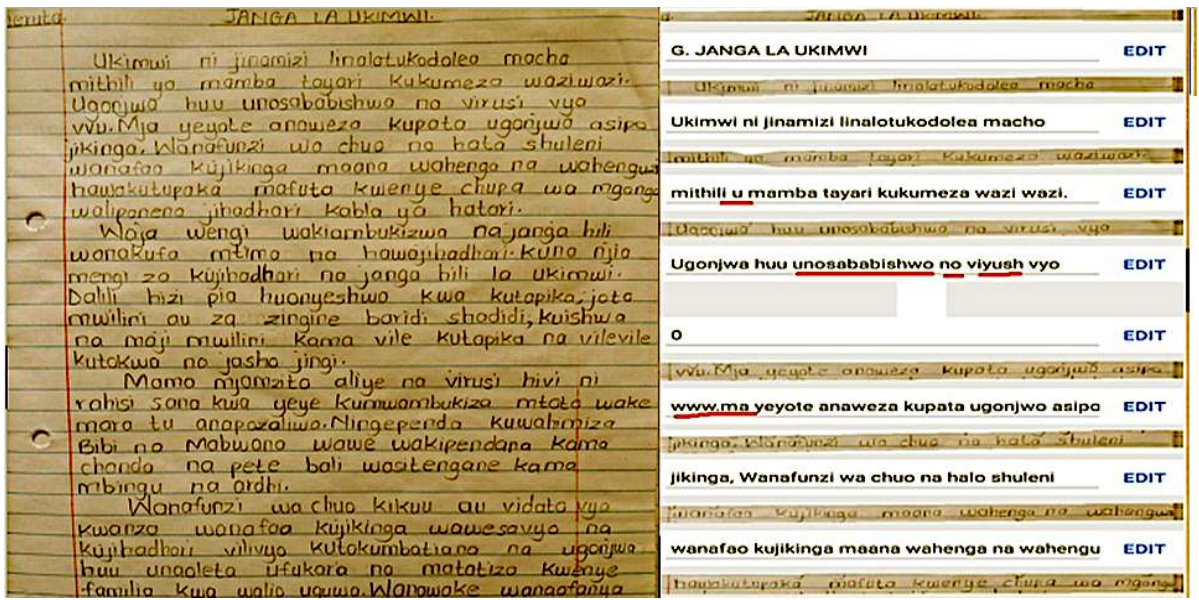

**Fig. 3** - Digitization using Pen to print app (source: author).

A comparative analysis of the various OCR tools confirmed that Google doc internal file converter had the best results. Therefore, Google doc internal file converter was extensively employed to convert the rest of the scanned files into editable text documents. The unique identifier of the data item was preserved in the naming of the cleaned texts to ensure direct linkage to the original source text. This internal file converter tool also reduced the project costs that would otherwise be used in procuring proprietary OCR software.

The second step in the data cleaning was the screening, which entailed analysis of the quantity and quality of the text output from the initial first step. This meant comparing the output of the OCR with the original hardcopy on a document-by-document basis. A sample of such a processing is shown on the right side of Fig. 3, where the original text is compared to the OCR converted text on a line-by-line basis.

Some of the original manuscripts had been improperly scanned. Such scans had resulted in images that were hazy and hence incomprehensible when digitized. To resolve this, the respective raw manuscripts were identified and rescanned properly. In some cases, the manuscripts contained crossed or strikethrough words that were incorrectly interpreted as ideographic characters by the OCR software. These were dropped while cleaning the final texts.

The whole raw text document was retyped in cases where the OCR output was non-intelligible.

The third data cleaning step was the diagnosis stage, where incomplete data, inconsistent punctuations, data redundancies, spelling variations, and other errors were identified, as shown in Fig. 4. An illustration of some errors that were noted in the typical OCR conversion are highlighted in the text on the right of Fig. 4, with reference to the original handwritten source text on the left.

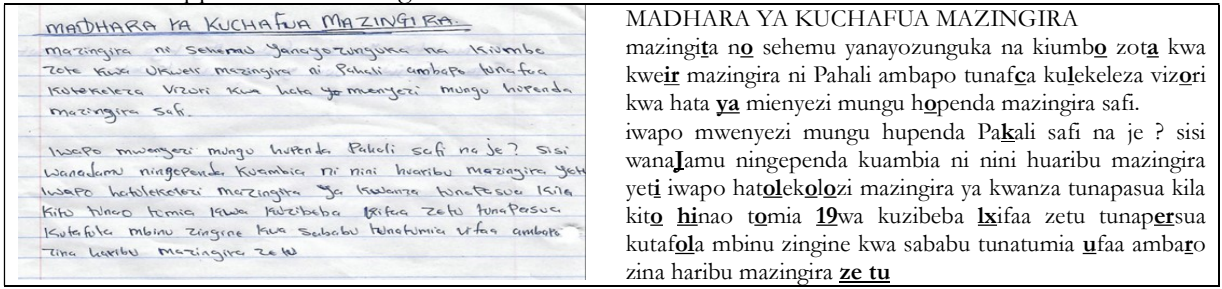

**Fig. 4** - Digitization of scanned PDF of Kencorpus storyID 1570 to an editable text document (source: author)

The fourth step involved treating the dataset by eliminating the errors identified in step 3. Corrections of any spelling, grammar and mistakes in sentence structure were also done in this stage of the cleaning. This was the longest, most time-consuming, and most expensive step. Note that the corrections (spelling, grammar, restructuring) were only done on the handwritten texts from schools. Based on the number of words, this was only 12% of the texts that required such treatment. Texts from media and publishers were retained as they were, with the processing only confirming that the final converted text was true to the original.

Fig. 5 below shows the typical processing of step 4, by first correcting the errors identified in step 3 (Fig. 4). The errors observed at this stage were the insertion of unknown characters or omission of characters from the original texts. The text on the left of Fig. 5 is exactly the output of step 3 as per Fig. 4 above, while the text on the right is the text resulting from corrections done on the identified errors that are shown in bold and underlined.

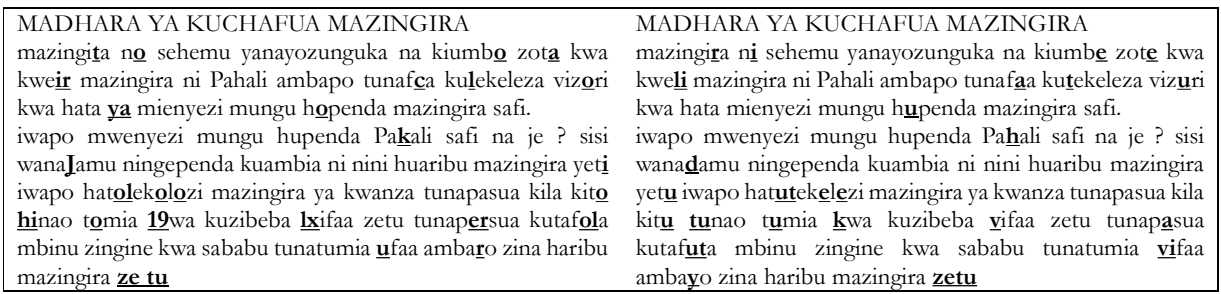

**Fig. 5** - Processing from Digitized to true to original TXT for part of Kencorpus storyID 1570 (source: author).

The second stage of step 4 involves any further corrections such as spelling, grammar, and sentence structures to ensure that the text is properly constructed to create meaning out of the original text. This is shown in Fig. 6 below, where introduction of proper nouns and other grammatical corrections have been made. The text to the left is the output of step 4, while the text on the right is the final reformatted text, considered as the final cleaned text from the process.

| MADHARA YA KUCHAFUA MAZINGIRA | MADHARA YA KUCHAFUA MAZINGIRA |
|---|---|
| mazingira ni sehemu yanayozunguka na kiumbe zote kwa kweli mazingira ni Pahali ambapo tunafaa kutekeleza vizuri kwa hata mienyezi mungu hupenda mazingira safi. iwapo mwenyezi mungu hupenda Pahali safi na je ? sisi wanadamu ningependa kuambia ni nini huaribu mazingira yetu iwapo hatutekelezi mazingira ya kwanza tunapasua kila kitu tunao tumia kwa kuzibeba vifaa zetu tunapasua kutafuta mbinu zingine kwa sababu tunatumia vifaa ambayo zina haribu mazingira zetu | Mazingira ni sehemu inanayozunguka viumbe vyote. Kwa kweli mazingira ni pahali ambapo tunafaa kutunza vizuri kwani hata Mwenyezi Mungu hupenda mazingira safi. Ikiwa Mwenyezi Mungu anapenda pahali pasafi na je sisi wanadamu? Ningependa kuelezea ni nini huharibu mazingira yetu tusipoitunza. Kuna vifaa vingi tunavyotumia ambavyo tukitupa ovyo ovyo huharibu mazingira yetu. Karatasi ni mojawapo ya vitu tunazotumia aambavyo huharibu mazingira yetu. |

**Fig. 6** - Processing from true to original TXT to final TXT for part of Kencorpus storyID 1570 (source: author).

The fifth and last step was the documentation stage. This involved indexing all of the original scanned files and mapping the 'level 0' master data with the respective cleaned data files stored in the 'level 1' cleaned data files. Subsequently, annotations based on the cleaned data could be done.

### 3.7 Data annotation

The project collected data for the Kencorpus corpus, and also developed datasets such as part of speech (POS) annotation for Dholuo and Luhya texts, QA pairs for Kiswahili texts, and translations from Dholuo to Kiswahili and Luhya to Kiswahili. We developed our own annotation guides for the POS tagging, translation, and QA tasks. These were made available to all research assistants for their reference. The details of all the data annotations aspects are pro-vided below.

### POS Annotation

Part of speech (POS) tagging was done by at least two research assistants per text document. The lead researcher in charge of the language did random checking on the POS tags and con-firmed agreement between the research assistants. The lead researcher, being the subject matter expert in the respective language, had the final decision in case of disagreement. POS tagging was based on a predefined tag set developed by the subject matter experts and shared amongst the annotators. This tag set was made uniform for deployment across the two languages being annotated (Dholuo and Luhya). The starting point of tag set development was the twelve universal tag set (Petrov et al., 2011). The initial idea was to increment the tags as necessary, upon consensus and agreement amongst the linguists in the four languages i.e., Dholuo and the three dialects of Luhya. This would result into a new tag set for uniform deployment across the corpus languages. However, the additional tags suggested for the different languages could not cut across all languages. After careful consideration, and to maintain uniformity across the languages in the project, a final decision was made to restrict our tagging to only the twelve universal tags. The deployed tagset is shown in Appendix 1.

The number of annotations done per language (Dholuo and Luhya) were one-third of the texts that had been digitized and cleaned. We did sampling based on the lengths of texts, targeting longer texts for the tagging.

### QA Annotation

Collection of Question-Answers pairs was done using an online form (Google form), where annotators filled in the QA pairs for each story text that they had read. The methodology for QA formulation was similar to what was employed for SQuAD (Rajpurkar et al., 2016) and TyDiQA (Clark et al., 2020). We formulated 5 QA pairs per text, with each annotator reading and setting the QA pairs as per predefined criteria such as the type of question to set and that the questions should be answerable by a single answer. The number of texts that were annotated with QA pairs was 70% of the Kiswahili texts. This volume of texts was chosen because it met the criteria set for doing the QA annotations. Such criteria spelt out the length of text (number of words) that could qualify for annotation. This was to avoid overly short or long texts to enable us to fit the standard number of 5QA pairs that we were to annotate on each text. Another criteria was on the text domain. This restricted our texts to prose and hence excluded poems and plays. The full details of the QA annotation aspects of the project are provided on a separate article (Wanjawa et al., 2023)

### Translation

The unit of translation adopted in the project was at sentence level. The annotators were to translate one sentence at a time. The translation method used was the literal translation of meaning of word(s) in context. We adopted equivalence translation, hence cultural words, idioms, metaphors, and sayings were translated into the best matching equivalence in the target language. At the beginning, the research had group translations where we conducted comparative translation to agree on the guidelines. Native speakers and linguistic experts participated in the translation. As supplementary material, dictionaries (Aswani (1995); Marlo et al., 2008; Ndanyi (2005); Odaga (1997); Parker (1998); Sibuor (2013); TUKI, 2013) were used as references. The translations done were for Dholuo-Kiswahili and Luhya-Kiswahili language pairs. The texts selected for translation were sampled through convenient sampling, targeting texts that were digitized and cleaned.

### 3.8 Quality control check

The project set up a system of checking each aspect of the project and ensuring that quality was assured. Quality control checking of all our processes from collection to annotation was a continuous process. This review assisted in ensuring that we provided datasets that were of high quality for use in machine learning tasks. We did the monitoring of data items as they were being collected and transmitted. This enabled the project to check on the data as it came and to apply corrective actions in terms of the quality of the data itself or the process of get-ting the data to the centralized storage. The collected data was submitted and managed in a shared centralized storage under the accountability of one researcher, who served as the data manager. The researcher in charge of raw data checked each received data item and communicated back to the research assistants in cases where the data needed to be resubmitted for whatever reason.

Data collection was supervised by the linguists, while the researchers also visited the data collection sites to confirm and spot check the data collection initiatives. We developed a processing and review methodology that called upon the research assistants to send data immediately upon collection instead of keeping it for long. In terms of project management and quality checks, each member of the research team was assigned a task and worked with a team of research assistants to accomplish the task. The research team met weekly to review, correct and guide on project execution to ensure the achievement of a quality product.

Data cleaning teams were guided on tools and methods of data cleaning, with the supervising researcher checking on the outputs. Each annotation work was done under a supervising researcher to ensure that the expected task was done, confirmed, and checked.

POS tagging was rechecked by the subject matter experts on a random sampling basis targeting 10% of all the annotated words. This was done by sampling the various genres of texts, but fully checking all the tags in the sampled text under consideration.

For translation, the researcher in charge, being the subject matter expert, also sampled 10% of the translated work and checked the translations to confirm accuracy. QA pair tagging task had a counterchecking system where the annotators reviewed a sample of each other's work to confirm agreement. The researcher in charge of QA annotation was to have the final decision in case of any disagreement on the set of QA pairs.

Lastly, the two proof-of-concept systems were overseen by the project team under the supervision of two researchers with expertise in the respective modeling (STT or QA). The teams and experts checked and confirmed that the project expectations of the proof-of-concept systems were met.

## 4 Results

The result of this project is the Kenyan languages corpus (Kencorpus). This is a corpus of texts and speech for the three languages of Kiswahili, Dholuo and Luhya. The project also developed other datasets of Dholuo-Kiswahili and Luhya-Kiswahili translations, part of speech (POS) tagged data for the Dholuo and Luhya languages and a question answering dataset for the Kiswahili language. Two proof of concept systems (STT and QA tasks) were also developed to test the datasets on NLP tasks.

### 4.1 Kecorpus statistics

The details of the dataset that the Kencorpus project collected is shown in Table 4.1 below. The final outcome of the initiative is therefore the Kenyan Languages corpus (Kencorpus) that has a dataset of 4,442 texts and 1,152 speech files. Note that the total numbers (Texts, Speech and Total) refer to the unique filenames compiled in the collection. Words refers to the total number of words in the texts available in the collection. The total time is the number of hours, minutes and seconds in speech files in the collection.

*Table 4.1: Kencorpus project statistics*

| Language | Texts | Words | Speech | Time | Total |
|---|---|---|---|---|---|
| Swahili | 2,585 | 1,829,727 | 104 | 19:10:57 | 2689 |
| Dholuo | 546 | 1,346,481 | 512 | 99:03:08 | 1058 |
| Luhya | 987 | 2,272,957 | 536 | 58:15:41 | 1523 |
| Tweets (Swahili) | 324 | 152,750 | 0 | 0:00:00 | 324 |
| Total | 4,442 | 5,601,915 | 1,152 | 176:29:46 | 5,594 |

The Luhya language data of a total of 1,523 data items as per Table 2, was however obtained from three different dialects, each dialect with its own data distribution as shown bro-ken down in Table 4.2 below.

*Table 4.2: Kencorpus project breakdown of Luhya language data into dialects*

| Dialect | Texts | Words | Speech | Time | Total |
|---|---|---|---|---|---|
| Luhya_Marachi | 483 | 67,812 | 138 | 15:37:46 | 621 |
| Luhya_Bukusu | 135 | 876,257 | 354 | 30:11:00 | 489 |
| Luhya_Logooli | 369 | 1,328,888 | 44 | 12:26:55 | 413 |
| Total | 987 | 2,272,957 | 536 | 58:15:41 | 1,523 |

### 4.2 Kecorpus annotations

The project created three datasets from the initial data collections above. The first dataset was a translation dataset of 13,400 sentences, being translations of Dholuo-Kiswahili and Luhya-Kiswahili sentence pairs as shown in Table. 4.3. The second dataset was a collection 143,000 words annotated for POS in two languages (Dholuo and Luhya-Lumarachi, Luhya-Lubukusu, Luhya-Lulogooli) as shown in Table 4.4. The final dataset was a dataset of 7,537 QA pairs from Kiswahili texts as shown in Table 4.5. More details about the Kencorpus Kiswahili Question Answering dataset (KenSwQuAD) including details of quality checks and machine learning systems developed as a result of that set are available as a separate project (https://doi.org/10.7910/DVN/OTL0LM).

*Table 4.3: Kencorpus datasets created from the data collected - Translation*

| Task | Sentences |
|---|---|
| Dholuo-Swahili translation | 1,500 |
| Luhya-Swahili translation | 11,900 |
| Total | 13,400 |

*Table 4.4: Kencorpus datasets created from the data collected – POS tagging*

| Task | Words |
|---|---|
| Dholuo POS tagging | 50,000 |
| Marachi POS tagging | 27,800 |
| Bukusu POS tagging | 30,900 |
| Logooli POS tagging | 34,300 |
| Total | 143,000 |

*Table 4.5: Kencorpus datasets created from the data collected – QA annotation*

| Aspect | No. |
|---|---|
| Swahili text stories annotated | 1,445 |
| Total QA pairs | 7,537 |

All the above-mentioned datasets and the corpus are available for use by researchers and any other user under the creative commons with attribution (CC BY 4.0) international license. These Kencorpus data collections are available on the project website (https://kencorpus.maseno.ac.ke/).

**4.3   Kencorpus proof of concept**

The Kencorpus project developed two proof of concept systems to test the datasets for their practical use in machine learning tasks. The first was to test the QA dataset on a QA system and the second was to test the collected speech data on a STT system. QA systems are applicable in systems such as internet search, dialog systems and chatbots, while STT systems are useful for processing text in instances such as educational materials for those with hearing challenges.

The first proof of concept that tested a QA system on the newly developed QA dataset was based on a deep learning system, specifically XLM-RoBERTa (Conneau & Lample, 2019). This was done using a dataset of 100 stories with 500 QA pairs from the Kencorpus QA dataset. The model used 80% of the data for training, while 20% of the data was used for testing. The file formatting and programming code for creating the model was based on what has been done in other similar settings for low-resource languages (Pytext, n.d.). The model settings were left at default apart from the following changes: Max sequence length = 384, Learning rate = 3e-5, batch size = 12, Epochs = 40 and max split span = 128.

The other machine learning method tried for QA was that of using semantic networks as al-ready tried on other datasets (Wanjawa & Muchemi, 2021). This method creates a network of nodes and edges based on the part of speech (POS) tagging of the text. The network then models the language by connecting the subjects-predicates-objects to create a network that shows the inter-relatedness of the POS words in the language. POS tagged words such as nouns tend to represent subjects and objects, while POS tagged words such as verbs tend to represent the relationships between the subjects and objects for SVO languages such as Swahili (Hurskainen, 2004a; Ndung'u, 2015). Such a network is created using programming scripts that mine the POS tagged text to identify the various parts of speech that represent subjects and objects, and then create the interconnections using POS tags such as verbs. The final format of the network is in a resource description format (RDF) which can then be queried using a database query language such as SPARQL, to enquire on the subjects or objects presented on the query text.

The second proof of concept model was a speech-to-text (STT) model based on the Swahili speech files collected in the project. This STT system was based on a collection of the speech corpus that is 27hrs 31min 50 sec, with 7 male and 19 female speakers. This created a Kiswahili phoneme dictionary of 31,759 word-phoneme pairs (Awino et al., 2022). The SST model was based on CMU Sphinx STT toolkit (Pantazoglou et al., 2018; Reddy et al., 2015).

The summary of performance of the two proof of concept systems developed for the project is shown in Table 4.6, with EM being exact match, while WER being word error rate. Details of the STT research are further explained in Awino et al. (2022) while those of the QA system are further explained in Wanjawa et al. (2023).

*Table 4.6: Kencorpus proof of concept system performance*

| Test system | Accuracy |
|---|---|
| QA system using deep learning (BERT) | 59.4% F1 score |
| QA system using semantic network | 80% EM |
| STT using CMU Sphinx STT toolkit | 18.87% WER |

While performance of QA systems for English language now uses machine learning to achieve upto F1 score of 93.214 (Paperswithcode, n.d.), the use of non-machine learning method for Kiswahili low resource language achieved 80% on exact match basis. The testing of the QA using a machine learning system (BERT) achieved an F1 score of 59.4%. While exact match and F1 metrics are used in different learning methods, the higher the performance the better the method. We observe that learning systems trained on little data available for low-resource languages are not able to perform as well as the models trained on large data such as for English, nor can such perform better than other system that are not based on training data, such as SN method.

For the STT system with Kiswahili language, the 18.87% WER is relatively good performance. High resource languages such as English have some of the best WERs, that are as low as 5.1% (Kurata et al., 2017). However, low resource languages have shown WERs of over 10%, such as 14.2% for Somali and 27.2% for Wolof language (Gauthier et al., 2016; Nimaan et al., 2006). Training the models on model data for similar low resource STT systems such as ours should improve the current WER rates.

## 5    Discussions

This project developed the Kenyan Languages corpus (Kencorpus) that has datasets in the three languages of Kiswahili, Dholuo and Luhya. For this research, we sampled three Luhya dialects, hence the final corpus has data in the five distinct languages of Swahili, Dholuo, Luhya-Lumarachi, Luhya-Lulogooli and Luhya-Lubukusu. These datasets were developed through data collection from both primary and secondary data sources by the project researchers. The Luhya-Lumarachi dialect was included to enable possible future research in the influence of neighbouring languages to a language of study. Lumarachi speakers are neighbours to the more populous Dholuo language speakers. This aspect was not addressed in this research but is a possible future linguistic study.

Analysis of the text data collected in this project showed that most texts were about 2,000 words in length on average. Our data on speech files indicated that most speech files were much less than 5 minutes each, especially from storytelling scenarios. The longer speech files were those from media houses where in some cases the audio clips provided could run up to an hour. These relative data sizes are useful in planning for project resources in similar corpus creation projects.

Getting data from primary sources such as education institutions and the community pro-vided the corpus with rich data that reflected diversity in the age groups from lower primary to college levels. Gender and cultural diversity were also manifested at data collection points. We discuss other pertinent issues encountered and the lessons learnt below.

**Raw data** - Data collected from lower grade schools were a bit illegible due to their writing habits and were made worse by their use of pencils for writing. Scanning or retyping handwrit-ten texts to create the computer formats needed in the corpus was difficult, with the discovery of such challenges coming up much later in the project after the data collection phase had ended. A lesson learnt from this was the need to use darker pens when collecting handwritten stories. This is not the usual conventional medium darkness (HB) pencil that are otherwise used in such school settings. The pen used should be much darker, and preferably an ink or biro pen where possible and as appropriate to the type and convenience of the respondent.

While it was fast and relatively easy to get secondary data from our collaborators, we still had challenges in processing some of the data that was still unclear e.g., scans or photographed images of newspaper clippings. Sometimes the shortcomings in the quality of the scans were related to the equipment used in the scanning or the expertise of the operator. We had instances where the scanned images would cut off the edges of the raw text documents. This made the reproduction of the original texts to be difficult. Some secondary data was also quite voluminous e.g., book scans or Bible sections. These were challenging to scan and to eventually reprocess into a computer format. We continually gave feedback to our researchers on the quality of the data being transmitted so that they could improve on any shortcomings over time.

Speech files did not have many challenges apart from the quality of recording that depended largely on the equipment. However, most of the collections (over 60%) were provided by project collaborators from media houses which were already in high quality studio recording format and in a suitable compression format such as MP3. Speech files tended to be big in size and sometimes the recording gadgets, such as smartphones, would run short of storage memory. Careful planning of work to anticipate the expected data volumes and preparing the recording hardware settings in advance assisted us in addressing such challenges. We also encouraged the research assistants to post their data to our central storage as soon as they could and immediately after the data collection exercise. This ensured that the recorders had more than adequate resources at any given time.

***Data cleaning*** – A data cleaning process was used to convert our raw documents into computer processable formats, which were TXT for all texts and WAV for all speech files. This process would lead to the creation of the corpus that is ready for machine learning tasks. Our project planned for the data cleaning to be done after the conclusion of data collection, so that the data to clean would be fully available for cleaning when the data cleaning exercise started. We however soon realized that the volume of data from the raw sources that needed cleaning was too much, despite the time allocated to data cleaning being a bit limited.

We also had other project tasks such as translations, POS tagging and QA annotation which needed the cleaned data and could therefore not start at their planned time as they needed the cleaned data. We overcame this by allowing parallel running of both the data cleaning and the annotations. In cases such as QA annotation, we allowed the annotators to use the raw data (images, scans, PDF) and annotate the QA pairs from such raw data sources. Future projects dealing with corpus creation can benefit by starting the data cleaning process as soon as data collection commences. These two processes should run in parallel to ensure that the workload involved in data cleaning is balanced throughout the project cycle and any dependent data processing tasks, such as annotations, are not delayed.

The decision on whether cleaning of texts should also include correction of spelling and grammar remains an issue of concern. Text data from lower-level schools tended to have many of such mistakes. In our case we only made corrections (spelling, grammar, sentence constructions as necessary) on all the handwritten texts that were collected from schools and left all other texts from publishers and media in their original form. Such corrections were however done on a small proportion of the data which turned out to be 12% of the total texts, based on the number of words.

The corpus therefore consists of text data that is of well-structured language, with correct spelling and grammar, either as originally sourced from publishers or media (88% of all texts, based on number of words) or corrected by our researchers when necessary (12%). The original collected text data before cleaning is also a good data source for research on text changes as influenced by data cleaning, optical character recognition (OCR) and image processing. Though the corpus publishes the cleaned TXT data format, it is possible to obtain the original source texts on request for purposes of such research.

***Data volume*** - There were challenges experienced with the proof-of-concept systems which could be attributed to the low volume of data in the corpus for machine learning methods that needed lots of data. The performance of the BERT-based proof of concept system on QA was low due to inadequate data. This BERT-based QA system was observed to improve in performance only when the data volume was increased. It is for this reason that we tested the SN method as another proof-of-concept, and this provided much better performance since it did not rely on vast amounts of data that was not available anyway, but just needed the original text to the POS tagged. The STT system that was tested as another proof of concept also did not have much data to train on since the Swahili speech files were relatively few, compared to the texts. Lack of training and testing data therefore remains the challenge with low resource language processing when subjected to machine learning methods that need such datasets.

Our Kencorpus project therefore contributes to resources of such languages as Kiswahili, Dholuo and Luhya. This should start placing them at the realm of machine learning systems so that users can access the many technological benefits of machine processing e.g., education materials for the physically challenged, internet search, chatbots, frequently asked questions (FAQs) that are developed from models such as STT and QA. These models need a corpus and the accompanying annotations, both of which the Kencorpus project provides.

As this and other data collection and annotation efforts continue, we expect that machine learning researchers targeting Swahili and other low resource languages, many of which are in Africa, shall start building resources for the benefit of the language users.

However, we also encountered some challenges while developing the Kencorpus datasets. The first was how to deal with incentives to respondents. This was a challenge while getting primary data where respondents expected or asked for compensation to provide the needed data. This had not been planned and would also not have fit into the budget of the project. We decided to forgo any such potential data sources. The other challenge was the Coronavirus disease of 2019 (COVID-19), a pandemic that caused curfews and restrictions in movement. We overcame such through good planning, virtual engagements and targeted one-off events that would enable us to gather as much information or input as possible, with minimal physical contact.

## 6      Conclusion

This research described the creation of Kencorpus, which is a corpus of 5,594 data items, both speech and text, for Swahili, Dholuo and Luhya. Kencorpus was created through the collection of primary and secondary data from education institutions, community, media houses and publishers. The research also reports on the data cleaning efforts that were needed to ensure that the datasets were in the expected computer format that would aid in further machine learning processes. The resultant collection in Kencorpus is the set of 4,442 text documents of 5.6 million words and 1,152 speech files of 176.5 hours across the three languages.

We have reported our POS tagging efforts for Dholuo, Luhya-Lumarachi, Luhya-Lulogooli and Luhya-Lubukusu. These tags are useful for language processing systems such as spelling and grammar checkers, hence such tools for our low resource languages can now be developed in our word processing programs. The paper has explained the development of parallel corpora for Dholuo-Swahili and Luhya-Swahili. This is useful in increasing parallel corpora for machine translation systems. A QA dataset of 7,537 QA pairs was also reported. This is useful in ma-chine learning systems for machine comprehension tasks and enquiry systems such as chatbots, frequently asked questions (FAQs) and even internet search using low resource languages. We developed an STT transcription proof of concept for Swahili which will contribute in research related to the development of speech technology tools.

By monitoring the data collection and dataset creation process at every stage, we ensured that the resulting corpus and datasets were of high quality and were also of practical use in typical machine learning systems as demonstrated by our proof-of-concept systems in STT and QA using deep learning and semantic networks.

For future work, this research corpus can be updated with new data and datasets over time. This shall further enrich it and make it even more useful in machine learning models that re-quire lots of training data.

## Acknowledgements

This research was made possible by funding from Meridian Institute's Lacuna Fund under grant no. 0393-S-001 which is a funder collaboration between The Rockefeller Foundation, Google.org, and Canada's International Development Research Centre.

We also acknowledge the inputs of the following research assistants for their contributions to the project:

*Dholuo team:* Beryl Odawa, Bildad Okebe, Immaculate Ochieng, Jackline Okello, Jonathan Onyango, Jotham Ondu, Jude Abade, Mary Muma, Mercy Oduoll, Wilfred Okoth

*Luhya team:* Belinda Oduor, Caren Nekesa, Chrispinus Waswa, Dorine Lugendo, Edwin Odhiambo, Evans Owino, Frankline Mwaro, Jamleck Lugwiri, Joseph Ambwere, Josephat Bwire, Joyline Ingasiani, Judith Awinja, Kints Mugoha, Mactilda Makana, Martin Mulwale, Mary Kibigo, Mary Masinde, Marystella Mbaya, Morris Salano, Phelisters Simiyu, Samwel Nyongesa, Stanely Kevogo, Tobias Shikuku, Vivian Alivitsa, Yonah Namatsi

*Swahili team:* Alice Muchemi, Benard Okal, Eric Magutu, Henry Masinde, Japheth Owiny, Khalid Kitito, Naomi Muthoni, Patrick Ndung'u, Phillip Lumwamu, Rose Nyaboke

*Data Cleaning team:* Led by Abiud Wekesa, Hesbon Watamba, Joyce Nekesa, Victor Orembe

*Systems team:* Ebbie Awino, Edwin Onyango, Geoffrey Ombui, Mark Njoroge

*Logistics team:* Job Ajiki, Kay Atieno, Leila Awuor, Odhiambo Oduke, Susan Makhanu.

**Appendix 1**

A1 – The 12 universal tags used in Kencorpus Part of Speech (POS) tagging (source: Petrov et al., 2011)
1. NOUN (nouns)
2. VERB (verbs)
3. ADJ (adjectives)
4. ADV (adverbs)
5. PRON (pronouns)
6. DET (determiners and articles)
7. ADP (prepositions and postpositions)
8. NUM (numerals)
9. CONJ (conjunctions)
10. PRT (particles)
11. '.' (punctuation marks)
12. X (a catch-all for other categories such as abbreviations or foreign words)